\documentclass[]{template}

\usepackage{amsmath,amssymb}
\usepackage{url}
\usepackage{array}
\usepackage{marvosym}
\newcommand{\emailsymbol}{\mbox{\scriptsize\Letter}}

\title{VideoEditor-MoE: Scaling Complex Video Editing with Mixture of Experts}

\newauthor{Xiao Cao$^{1,2}$, Fuchen Long$^{1}$, Yansong Qu$^{3}$, Wen Xiao$^{2}$, Xin Zhang$^{2}$, Xinlong Sun$^{1}$, Xi Chen$^{1,\emailsymbol}$, Zhiyong Huang$^{2,\emailsymbol}$}
\affiliation[1]{\scriptsize{Tencent}}
\affiliation[2]{\scriptsize{National University of Singapore}}
\affiliation[3]{\scriptsize{Xiamen University}}
\affiliation[\emailsymbol]{\scriptsize{Corresponding Authors}}

\abstract{
Complex video editing requires understanding complex instructions while jointly fulfilling multiple editing requests. Meeting these demands through dense model scaling substantially increases computational cost. We introduce VideoEditor-MoE, the first open-source mixture-of-experts framework for video editing. Beyond sparse MoE-MLP layers, MoE-Attention expands attention capacity through routed key--value experts while preserving global interactions among video tokens. Initialized from a 5B video editing model, VideoEditor-MoE scales to 14B and 56B parameters, with approximately 8B parameters activated per token in both variants. Comparisons with existing methods and two complementary baselines, Explicit-MoE and MoE-MLP, assess its benefits over explicit expert guidance and MLP-only expert architectures. Further analyses demonstrate that all experts remain active and reveal that learned routing responds more strongly to sample content and diffusion noise levels than to human-defined editing categories. Spatial analysis further indicates sensitivity to object-level semantic features. We also diagnose asymmetric expert erosion under sparse gradients during 56B scaling.
}

\github{\url{https://github.com/massyzs/Video-Editor-MoE}}

\begin{document}
\thispagestyle{firstheader}
\maketitle
\pagestyle{empty}

\section{Introduction}
A complex video editing request succeeds only when every requested change is realized on the targeted objects. Recent models support diverse editing functionality, such as object addition, removal, style transfer~\citep{kiwiedit,univideo,teleomni,vace}. Yet, a real request often combines several operations across objects or applies multiple changes to the same object. 
\textit{``Turning a car blue while replacing the sky with a sunset''}, for example, requires both edits to persist as the scene evolves. A visually plausible result can still leave the request incomplete if one change is omitted or only weakly expressed. Complex editing therefore demands both reliable instruction coverage and complete visual modification, together with preservation of unrelated content and temporal coherence.

Increasing model capacity is a natural direction for supporting these complex instructions. Yet, scaling a dense video model increases computation throughout the network. Mixture-of-experts (MoE) architectures offer a way to expand model capacity while activating only a small subset of experts for each token~\citep{mixtral_moe,deepseekmoe}. To this end, scaling video editing models through MoE provides a practical approach to balancing computational cost and model capacity. However, conventional MoE-MLP expands token-wise transformations while leaving the attention architecture unchanged. Complex editing also requires the exchange of semantic information across regions and frames to coordinate multiple modifications. This motivates extending expert capacity to attention, allowing MoE semantic representations to participate in global token interactions and support coherent editing across space and time.



We introduce VideoEditor-MoE, the first open-source MoE framework for video editing, supporting complex instructions with multiple editing requirements. VideoEditor-MoE incorporates both MoE-MLP and MoE-Attention to expand model capacity beyond conventional dense architectures. MoE-Attention combines shared queries with routed, paired key--value experts. Each token contributes multiple key--value representations, and every query attends to the routed key--value pairs from all tokens. This design expands attention capacity while preserving global token interactions. Together with sparse MLP layers, we develop two model variants, 14B-A8B and 56B-A8B, both with approximately 8B activated DiT parameters per token.

In addition, two baselines are developed to provide complementary comparisons. Explicitly Guided MoE selects experts according to predefined editing categories that reflect an intuitive human understanding of editing tasks, testing whether such task divisions align with effective expert allocation. MoE-MLP expands capacity through MLP experts alone, serving as an ablation baseline for assessing the contribution of MoE-Attention. 
Beyond editing performance, we conduct detailed analyses of expert utilization, routing behavior, and scaling stability. All experts remain active during inference, and learned expert allocation responds more strongly to sample content and diffusion noise states than to human-defined editing categories. To further understand the content dependence, spatial analysis examines which visual features attract expert routing responses and suggests sensitivity to object-level semantic features. Finally, scaling diagnostics trace suppressed MoE outputs at 56B scale to asymmetric pretrained expert erosion under sparse task gradients, showing why decoupling weight decay prevents branch collapse.

Our contributions are:
\begin{itemize}
    \item We highlight complex instruction-based video editing as a practical setting that requires jointly satisfying multiple modifications to a single object or coordinating edits across multiple objects, motivating greater model capacity for compositional editing.
    \item We introduce VideoEditor-MoE, extending conventional MoE-MLP with MoE-Attention to expand sparse expert capacity into attention. This design expands model capacity and enables a more comprehensive MoE architecture for video generation.
    \item We demonstrate that all experts remain active and reveal that learned routing responds more strongly to sample content and diffusion noise levels than to human-defined editing categories. Spatial analysis further indicates sensitivity to object-level semantic features.
    \item We will publicly release both VideoEditor-MoE variants, 14B-A8B and 56B-A8B, together with two complementary baselines, to facilitate further research on MoE architectures for video generation and editing.
\end{itemize}

\section{Related Work}

\subsection{Instruction based Video Editing}

Instruction-based video editing aims to modify a source video according to natural-language instructions. Recent methods~\citep{minimaxh3,teleomni,univideo,kiwiedit,vace} have demonstrated promising editing results, using either text instructions alone or additional guidance such as masks and reference inputs. These advances enable a range of editing operations, but handling complex instructions that combine multiple editing requirements remains challenging.

In real-world scenarios, users may request edits to multiple objects, multiple modifications to a single object, or a combination of both. Such instructions require models to associate each modification with the correct target and coordinate multiple edits while maintaining temporal consistency and preserving unrelated content. Existing methods may omit some requested edits or only partially realize the intended changes. Meeting these demands places greater requirements on model capacity, motivating the exploration of larger video editing models. However, scaling dense models entails substantial increases in computational cost. This motivates us to explore sparse MoE architectures as a means of expanding model capacity while limiting the growth in computation cost.

\subsection{Mixture-of-Experts}

\subsubsection{Preliminary} 
Mixture-of-experts (MoE) architectures enable model capacity to scale through sparse expert activation. Two representative designs in large language models are Mixtral~\citep{mixtral_moe} and DeepSeekMoE~\citep{deepseekmoe}. Mixtral replaces dense feed-forward networks (FFNs) with multiple expert networks and a token-level router. Given an input token representation $x$, the MoE module computes
\begin{equation}
    F_{\mathrm{Mixtral}}(x)
    =
    \sum_{i=0}^{n-1} G(x)_i\,E_i(x),
\end{equation}
where $n$ is the number of experts, $E_i$ denotes the $i$-th expert, and $G(x)_i$ is its routing weight. The router selects the top-$k$ experts for each token and assigns zero weights to the remaining experts.

DeepSeekMoE introduces fine-grained expert segmentation and shared expert isolation to improve parameter efficiency. Specifically, it reduces each expert's FFN intermediate dimension by a factor of $m$, yielding $mn$ smaller experts, and designates $k_s$ of them as shared experts that process every token. Using the same notation for expert outputs and sparse routing weights, its MoE module can be expressed as
\begin{equation}
    F_{\mathrm{DeepSeek}}(x)
    =
    \sum_{i=0}^{k_s-1} E_i(x)
    +
    \sum_{i=k_s}^{mn-1} G(x)_i\,E_i(x),
\end{equation}
where the router selects $mk-k_s$ experts from the $mn-k_s$ routed experts. Together with the $k_s$ shared experts, this configuration activates $mk$ fine-grained experts per token, preserving approximately the same expert computation budget as activating $k$ original-sized experts. Fine-grained segmentation enables more flexible expert combinations, while shared experts aim to capture common knowledge and reduce redundancy among routed experts. Here, $G(x)_i$ denotes the routing weight under each architecture's gating rule, and both formulations omit residual connections.

\subsubsection{Related Methods}

MoE architectures have been explored in visual generation to expand model capacity and adapt computation to different inputs and denoising stages. DiT-MoE~\citep{dit_moe} introduces shared and routed MLP experts into diffusion transformers, increasing total capacity while limiting the computation activated for each token. Switch-DiT~\citep{switch_dit} uses timestep-conditioned routing to mitigate interference between denoising tasks at different noise levels. Diff-MoE~\citep{diff_moe} combines expert-specific timestep conditioning with spatially adaptive expert allocation, allowing expert computation to adapt to both token content and the generative stage. 

MoE has also been explored in video generation. Wan2.2~\citep{wan22} assigns high- and low-noise denoising stages to two specialized dense experts, separating coarse content formation from detail refinement. This design allocates computation at the model level according to the noise stage. Mamoda2.5~\citep{mamoda25} instead incorporates fine-grained routed experts and a shared expert into a unified multimodal framework, expanding capacity for understanding, video generation, and editing with sparse activation, though it has not yet been publicly released. LingBot-Video~\citep{lingbot_video} explores sparse MoE scaling for embodied video pretraining, targeting physical dynamics and action-conditioned generation.

These advances motivate an open-source MoE framework tailored to the demands of complex video editing, where multiple editing requirements call for greater model capacity. We introduce the first open-source MoE-based video editing model, with pretrained video editing models providing a practical starting point for scaling. A Mixtral-style MoE design enables expert initialization from existing models, while MoE-Attention further expands attention capacity to support interactions among video tokens. Together, these designs offer a practical path to scaling complex video editing capabilities while retaining the knowledge acquired by pretrained models.

\section{Method}
In this section, we first introduce two additional baselines: (1) Explicitly Guided MoE and (2) MoE-MLP, to examine MoE routing preferences and demonstrate the benefits of our MoE-Attention design, respectively. We then present the design of VideoEditor-MoE. All the baselines and our proposed VideoEditor-MoE are built on the same 5B 30-layer base model.

\subsection{Baselines}

\paragraph{Explicitly Guided MoE.}
This baseline guides expert selection using five categories that reflect an intuitive human understanding of editing tasks: object addition, object removal, object replacement, style transfer, and background replacement. As shown on left side of Appendix Figure~\ref{fig:baseline}, a main editing branch integrates features from task-specific experts, each fine-tuned on one category with low-rank adaptation (LoRA)~\citep{lora}.

A VLM~\citep{qwen,qwen3} decomposes the complex instruction into individual edits in advance. The main branch receives the complete instruction, while each expert receives the instruction for its assigned edit. A router, pretrained on an editing-category classification task, predicts each individual edit's category from VLM features. Fusion layers are inserted into the first half of the main branch. These layers perform semi-attention: main-branch queries attend to main-branch tokens and expert key--value features, with spatial offsets applied to the RoPE~\citep{rope} embeddings of the expert features.

\paragraph{MoE-MLP Baseline.}
This baseline processes the complete instruction without editing-category supervision. It expands MLP capacity while keeping attention dense, serving as an ablation baseline for MoE-Attention. For the DiT side, MLPs at even blocks $\mathcal{M}=\{2,4,\ldots,28\}$ are replaced with MoE layers while other blocks remain dense. Each MoE layer contains $N$ full-width MLP experts and a token-level router, with no shared expert. For MoE blocks, a video-token $x$ is represented by:
\begin{equation}
    p(x)=\operatorname{softmax}(xW_g),
    \qquad
    G(x)_i=\frac{p(x)_i\,\mathbf{1}[i\in\mathcal{I}_k(x)]}{\sum_{j\in\mathcal{I}_k(x)}p(x)_j},
    \qquad
    F_{\mathrm{MoE}}(x)=\sum_{i=1}^{N}G(x)_iE_i(x),
    \label{eq:moe_routing}
\end{equation}
where $W_g$ is the router matrix, $\mathcal{I}_k(x)$ contains the top-$k$ expert indices, and $\mathbf{1}$ is the indicator function. The selected weights $G(x)_i$ sum to one, while unselected experts receive zero weight. The final output combines the selected MLP expert outputs $E_i(x)$. The overall framework is shown in Appendix Figure~\ref{fig:baseline} right side. We set $N=12$ and choose top-3 routing, yielding a \textbf{19B-A8B} MoE-MLP baseline.


\subsection{VideoEditor-MoE with MoE-Attention}
\begin{figure}[t]
    \centering
    \includegraphics[width=\linewidth]{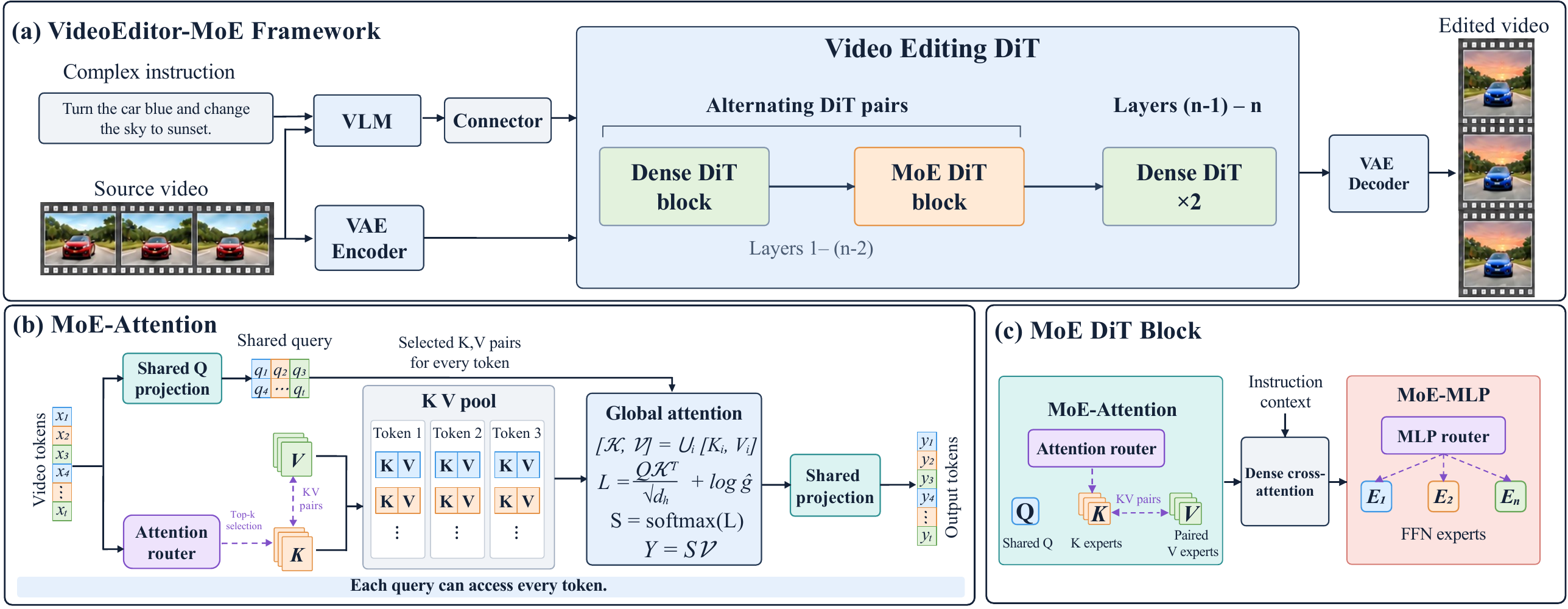}
    \caption{\textbf{VideoEditor-MoE.} (a) Dense and MoE DiT blocks alternate through 28 layers, followed by two dense blocks. (b) Shared queries attend to routed key--value pairs from all video tokens. (c) Each MoE block combines MoE-Attention, dense instruction cross-attention, and MoE-MLP.}
    \label{fig:framework}
\end{figure}

MoE-MLP transforms individual tokens, while attention exchanges semantic information across regions and frames. VideoEditor-MoE extends the MoE-MLP baseline with paired key--value experts in self-attention mechanism within MoE blocks. As shown in Figure~\ref{fig:framework}(b), query and output projections remain shared. Attention and MLP experts use separate routers, and instruction cross-attention remains dense.

For $T$ video-token representations $X\in\mathbb{R}^{T\times d}$, the attention router applies the top-$k$ rule in Equation~\ref{eq:moe_routing}. Each selected index $i$ chooses both a key projection $W_K^{(i)}$ and its paired value projection $W_V^{(i)}$. Let $X_i$ contain the tokens that select expert $i$; a token appears in $k$ such groups. For one attention head of dimension $d_h$,
\begin{equation}
    Q=XW_Q,\qquad
    K_i=X_iW_K^{(i)},\qquad V_i=X_iW_V^{(i)},
    \label{eq:expert_kv}
\end{equation}
where $W_Q,W_K^{(i)},W_V^{(i)}\in\mathbb{R}^{d\times d_h}$. Biases, query/key normalization, and RoPE are omitted from the equations for clarity. They follow the base model, and each key representation retains its source token's position.

Following Figure~\ref{fig:framework}(b), all selected key--value pairs are written as
\begin{equation}
    [\mathcal{K},\mathcal{V}]=\bigcup_i[K_i,V_i],
    \qquad \mathcal{K},\mathcal{V}\in\mathbb{R}^{kT\times d_h}.
    \label{eq:routed_kv}
\end{equation}
Here, $\bigcup$ collects the selected key--value pairs from all tokens, keeping each key together with its corresponding value. Each token contributes $k$ pairs, giving $kT$ pairs in total. The pair produced by expert $i$ for token $x_t$ carries the normalized routing weight $G^{\mathrm{attn}}(x_t)_i$, computed using Equation~\ref{eq:moe_routing}. These weights form the vector $\hat g\in\mathbb{R}^{1\times kT}$, ordered to match the corresponding pairs in $\mathcal{K},\mathcal{V}$.
Every query attends to all $kT$ routed pairs through a single attention operation:
\begin{equation}
    L=\frac{Q\mathcal{K}^{\top}}{\sqrt{d_h}}+\log\hat g,
    \qquad S=\operatorname{softmax}(L),
    \qquad Y=S\mathcal{V}.
    \label{eq:moe_attention_output}
\end{equation}
The term $\log\hat g$ adds the logarithm of each pair's routing weight to every query's attention scores. It favors pairs preferred by the router, while $Q\mathcal{K}^{\top}$ measures their relevance to each query. Softmax normalizes each row over all $kT$ pairs, and $Y$ combines their values using these attention weights. The head outputs pass through the shared output projection.

MoE-Attention produces multiple expert-specific key--value representations for each token, enabling richer semantic modeling. Every query attends to the routed key--value pairs from all tokens, preserving global interactions across regions and frames.


\paragraph{Training objective.}
Training combines flow-matching~\citep{flow_matching} reconstruction, temporal-difference supervision, and per-sample expert load balancing as in Appendices~\ref{app:loss} and~\ref{app:implementation}.

\section{Experiments and Analysis}
We first compare edit execution, fulfillment, output quality, and FVD~\citep{fvd} over three benchmarks: (1) OpenVE-Bench~\citep{openve}, (2) CoinVE-Bench~\citep{coinve}, and (3) CoVE-Bench~\citep{covebench}. We then examine (1) \textbf{computational cost}, (2) \textbf{expert utilization}, and (3) \textbf{routing sensitivity} to diffusion noise levels and sample content. Building on this content sensitivity, we further investigate (4) \textbf{spatial object tracking}: whether content-driven routing tracks moving semantic objects rather than static spatial or color priors. Finally, we diagnose (5) \textbf{asymmetric expert erosion} under sparse task gradients when scaling the expert pool to 56B and why it suppresses MoE branches.

\begin{table}[t]
    \centering
    \caption{Quantitative comparison. VideoEditor-MoE-56B-A8B ranks first and 14B-A8B ranks second in edit execution rate, edit fulfillment, and video quality, while maintaining competitive FVD. We show averaged scores over three benchmarks here and detailed scores in supplementary material. Evaluation protocol appears in Appendix~\ref{app:selected_case_protocol}.}
    \label{tab:selected_case_results}
    \small
    \setlength{\tabcolsep}{3pt}
    \begin{tabular*}{\linewidth}{@{\extracolsep{\fill}}l cccc cc cc@{}}
        \toprule
        & \multicolumn{4}{c}{Existing methods} & \multicolumn{2}{c}{MoE baselines} & \multicolumn{2}{c}{Ours} \\
        \cmidrule(lr){2-5} \cmidrule(lr){6-7} \cmidrule(l){8-9}
        Metric & Kiwi-Edit & \begin{tabular}{@{}c@{}}Kiwi-Edit\\(Finetuned)\end{tabular} & \begin{tabular}{@{}c@{}}MiniMax-\\H3\end{tabular} & \begin{tabular}{@{}c@{}}CoinVE-\\Edit\end{tabular} & \begin{tabular}{@{}c@{}}Explicit-\\MoE\end{tabular} & \begin{tabular}{@{}c@{}}MoE-MLP\\(19B-A8B)\end{tabular} & 14B-A8B & 56B-A8B \\
        \midrule
        Edit Execution Rate $\uparrow$ & 0.678 & 0.706 & 0.736 & 0.779 & 0.724 & 0.810 & 0.837 & \textbf{0.853} \\
        Edit Fulfillment $\uparrow$ & 0.675 & 0.694 & 0.727 & 0.778 & 0.727 & 0.813 & 0.834 & \textbf{0.851} \\
        Video Quality $\uparrow$ & 0.592 & 0.576 & 0.598 & 0.677 & 0.650 & 0.683 & 0.704 & \textbf{0.727} \\
        FVD/1000 $\downarrow$ & 1.213 & 1.152 & 1.493 & 1.095 & 1.099 & 1.268 & 1.193 & \textbf{1.025} \\
        \bottomrule
    \end{tabular*}
\end{table}

\subsection{Quantitative Results}
\label{sec:selected_case_results}
We evaluate edited videos along four dimensions. \textbf{Edit Execution Rate} measures the fraction of requested edits successfully performed; \textbf{Edit Fulfillment} measures how completely and accurately each change is realized, including unsuccessful edits; \textbf{Video Quality} assesses visual quality, temporal consistency, preservation of unedited content, and integration of the edits; and \textbf{Fréchet Video Distance (FVD)} assesses video realism and temporal coherence by measuring the distributional distance between edited and reference videos. The evaluations are conducted on three benchmarks: (1) OpenVE-Bench, (2) CoinVE-Bench, and (3) CoVE-Bench. The averaged scores over three benchmarks are shown in Table~\ref{tab:selected_case_results}, and detailed scores over each benchmark can be found in supplementary material. The evaluation protocol and prompt templates are provided in Appendix~\ref{app:selected_case_protocol} and Figures~\ref{fig:prompt1}, \ref{fig:prompt2}, and~\ref{fig:prompt3}.

As shown in Table~\ref{tab:selected_case_results}, VideoEditor-MoE-56B-A8B achieves the highest scores across all metrics, followed by the 14B-A8B variant. Both variants outperform the six baselines in edit execution, edit fulfillment, and video quality. As an ablation baseline, MoE-MLP baseline ranks third on both editing metrics, while its gap to VideoEditor-MoE highlights the importance of MoE-Attention.

\begin{figure}[t]
    \centering
    \includegraphics[pagebox=cropbox,width=\linewidth]{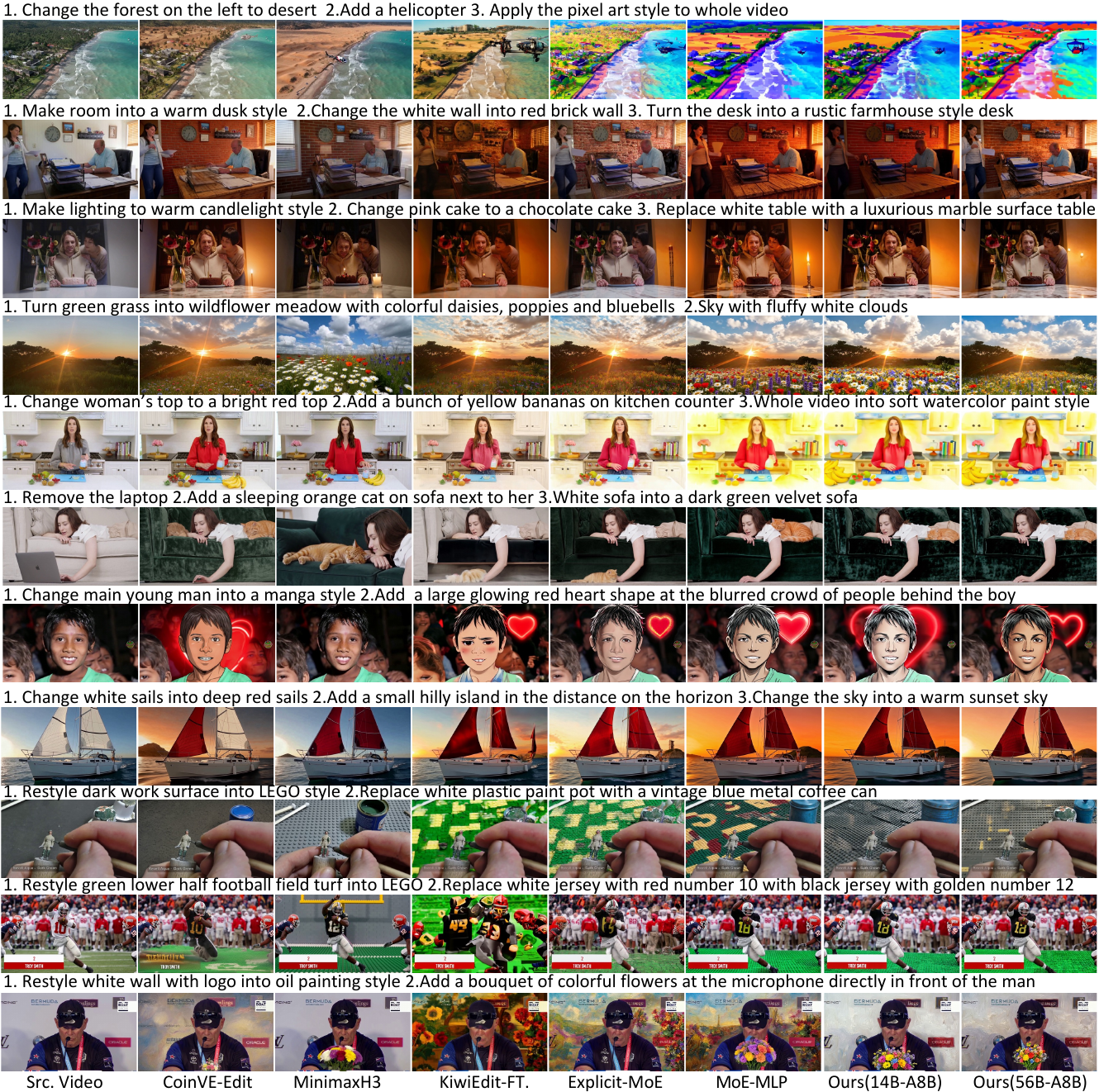}
    \caption{\textbf{Qualitative comparisons under complex instructions.} Each row shows the source video and outputs from five baselines and both VideoEditor-MoE variants. Middle frames are shown, with editing instructions summarized above each row.}
    \label{fig:visual}
\end{figure}

\subsection{Qualitative Results}
\label{sec:qualitative}
As shown in Figure~\ref{fig:visual}, we show both VideoEditor-MoE variants with five of six baselines due to space constraints (excluding the original KiwiEdit). For compact presentation, only the middle frame of each video is shown, and the displayed instructions are manually decomposed and simplified for readability.

VideoEditor-MoE jointly realizes multiple requested changes within the same scene. In the third row of Figure~\ref{fig:visual}, our variants capture the requested candlelit atmosphere through warm scene lighting, while some baselines emphasize inserting visible candles. In the fifth row, both variants combine soft watercolor stylization with the addition of bananas. In the sixth row, VideoEditor-MoE and MoE-MLP baseline place the cat beside the woman while preserving the scene structure. Most baselines output omit requested changes or express them less clearly, illustrating the challenge of jointly fulfilling complex instructions.

\subsection{Computational Cost}

Table~\ref{tab:computational_cost} reports the DiT TFLOPs for one denoising step and DiT time cost of generating one video. With optimized inference, the computation of VideoEditor-MoE scales approximately in proportion to its activated rather than total parameters. Both variants activate approximately 8B parameters per token and require 95 TFLOPs, so expanding the expert pool from 14B to 56B leaves TFLOPs unchanged. Compared with MiniMax-H3 (24B) and CoinVE-Edit (14B), VideoEditor-MoE-56B-A8B requires $22$--$31\times$ fewer TFLOPs and $23$--$31\times$ less time, while achieving better editing performance (Table~\ref{tab:selected_case_results}). These results demonstrate the efficiency and practicality of VideoEditor-MoE.

\begin{table}[t]
    \centering
    \caption{\textbf{Computational cost}. VideoEditor-MoE-56B-A8B requires far fewer TFLOPs and less time than MiniMax-H3 (24B) and CoinVE-Edit (14B). TFLOPs are reported for a single DiT denoising step, and DiT time cost is for generating one video.}
    \label{tab:computational_cost}
    \small
    \setlength{\tabcolsep}{3pt}
    \begin{tabular*}{\linewidth}{@{\extracolsep{\fill}}l cccc cc cc@{}}
        \toprule
        & \multicolumn{4}{c}{Existing methods} & \multicolumn{2}{c}{MoE baselines} & \multicolumn{2}{c}{Ours} \\
        \cmidrule(lr){2-5} \cmidrule(lr){6-7} \cmidrule(l){8-9}
        Cost & Kiwi-Edit & \begin{tabular}{@{}c@{}}Kiwi-Edit\\(Finetuned)\end{tabular} & \begin{tabular}{@{}c@{}}MiniMax-\\H3\end{tabular} & \begin{tabular}{@{}c@{}}CoinVE-\\Edit\end{tabular} & \begin{tabular}{@{}c@{}}Explicit-\\MoE\end{tabular} & \begin{tabular}{@{}c@{}}MoE-MLP\\(19B-A8B)\end{tabular} & 14B-A8B & 56B-A8B \\
        \midrule
        TFLOPs        & 55  & 55  & 2090 & 2984 & 260  & 80   & 95   & 95   \\
        Time cost (s) & 16.4 & 17.4 & 596 & 806 & 99.2 & 40.2 & 16.5 & 26.1 \\
        \bottomrule
    \end{tabular*}
\end{table}

\subsection{Expert Analysis}
\label{sec:expert_analysis}

We analyze expert utilization and input-dependent routing on the 14B-A8B variant across complex editing inputs and denoising noise levels $\sigma\in\{0.355,0.833,0.978\}$, with protocol details in Appendix~\ref{app:expert_comparisons}.

\begin{figure}[t]
    \centering
    \includegraphics[width=\linewidth]{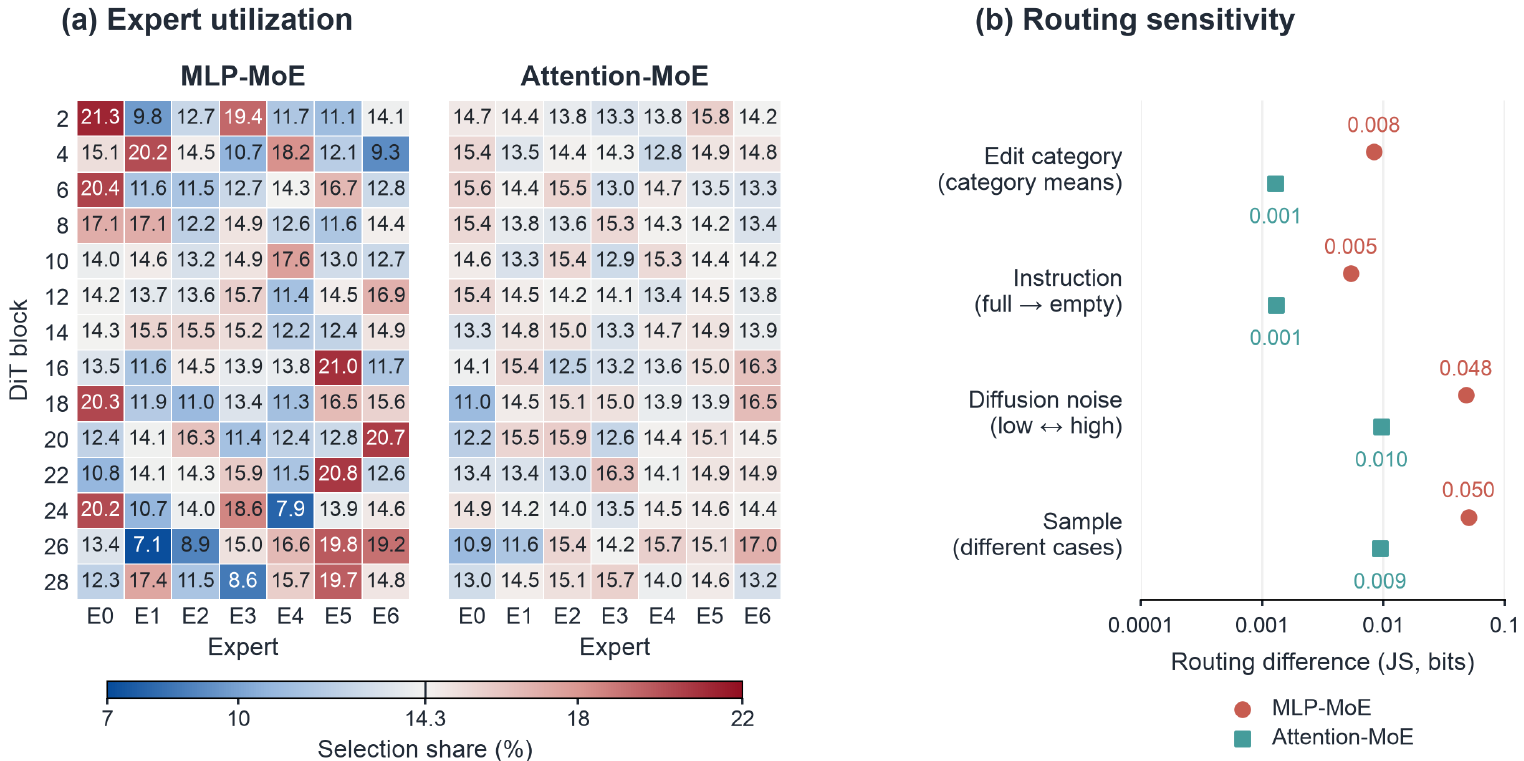}
    \caption{\textbf{Expert utilization and input-dependent allocation.} (a) Mean expert selection shares across evaluated videos and noise levels; each row sums to $100\%$ (uniform: $14.3\%$). (b) Mean video-level Jensen--Shannon (JS) divergence across controlled and cross-sample comparisons. Larger values indicate stronger global expert reallocation.}
    \label{fig:expert_analysis}
\end{figure}

\begin{figure}[t]
    \centering
    \includegraphics[width=\linewidth]{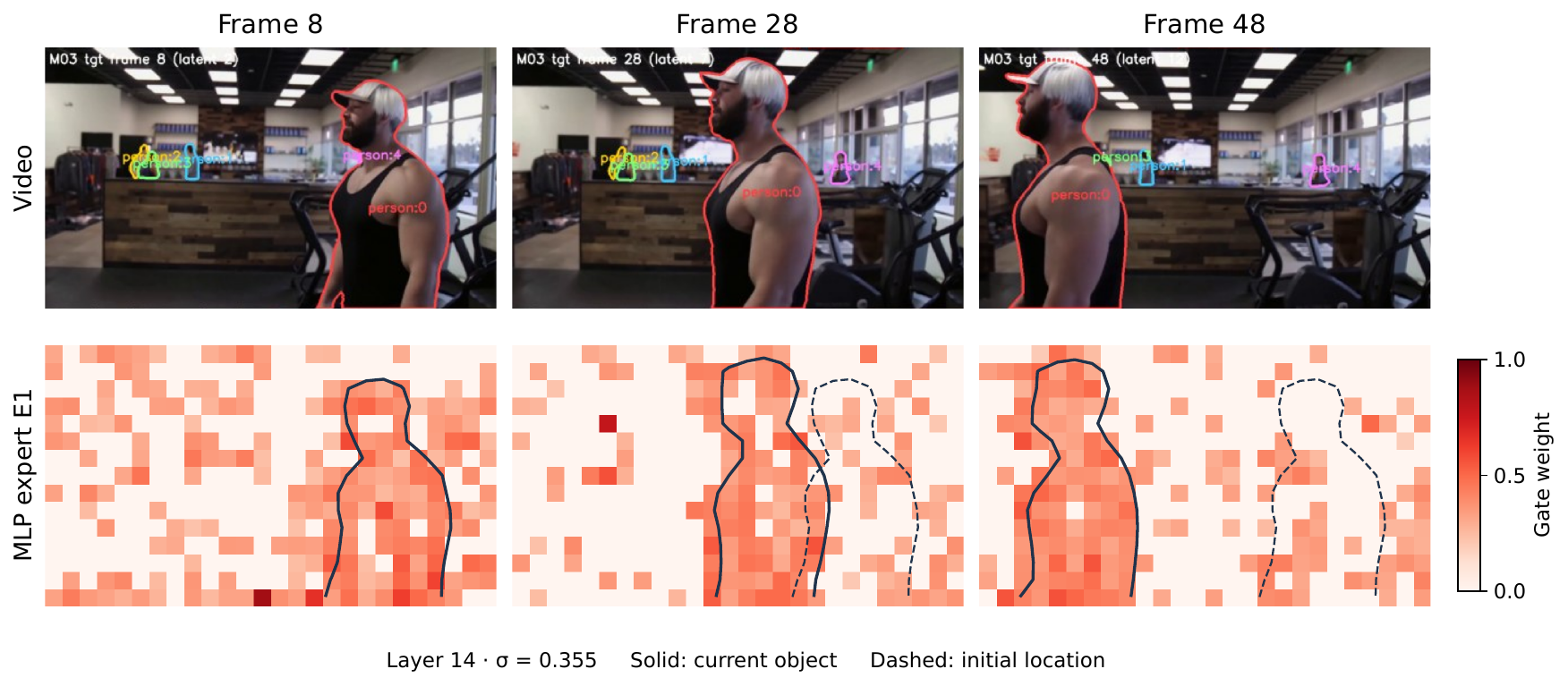}
    \caption{\textbf{Object-semantic expert routing.} Top: source frames (red contour marks the tracked person mask). Bottom: gate weights of a representative MLP expert (Layer 14, $\sigma=0.355$), which track the moving subject and recede at its initial location (dashed).}
    \label{fig:expert_motion}
\end{figure}

\paragraph{All experts remain active.}
Each of $T$ video tokens selects $k$ experts, yielding $kT$ selections. Expert $i$'s video-level selection share is $s_i=\frac{\sum_{t=1}^{T}\mathbf{1}[i\in\mathcal{I}_k(x_t)]}{kT}$, where $\mathcal{I}_k(x_t)$ contains the experts selected for token $x_t$. Summarizing sequence-level expert budgets directly tests whether individual experts act as global task-category modules as posited by Explicitly Guided MoE. As shown in Figure~\ref{fig:expert_analysis}(a), even the least-used experts receive mean shares of $7.1\%$ in MoE-MLP and $10.9\%$ in MoE-Attention, and \textbf{all 196 layer--router--expert units remain active across every evaluated setting}, confirming that no expert collapses.

\paragraph{Routing is more sensitive to content and noise level.}
We evaluate sequence-level routing sensitivity across editing categories, instructions, diffusion noise levels, and video content in Figure~\ref{fig:expert_analysis}(b). For two conditions or videos $a$ and $b$ at the same layer and router, we form $\mathbf p=[s_1^{(a)},\ldots,s_N^{(a)}]$ and $\mathbf q=[s_1^{(b)},\ldots,s_N^{(b)}]$ and measure their Jensen--Shannon (JS) divergence~\citep{js} in bits:
\begin{equation}
    \operatorname{JS}(\mathbf p,\mathbf q)=\tfrac12\sum_i p_i\log_2\frac{p_i}{m_i}+\tfrac12\sum_i q_i\log_2\frac{q_i}{m_i},
    \qquad \mathbf m=(\mathbf p+\mathbf q)/2.
    \label{eq:js}
\end{equation}
Because averaging within categories smooths sample variance, our primary controlled test uses unaveraged, matched per-video interventions that fix the video and noise realization while toggling full versus empty instructions or low versus high $\sigma$. We also compute pairwise JS across distinct videos and across the five editing-category mean profiles, averaged over all 14 MoE layers separately for MLP and attention (Appendix~\ref{app:expert_comparisons}).

In Figure~\ref{fig:expert_analysis}(b), under matched per-video controls without group averaging, shifting $\sigma$ alters MoE-MLP allocation nearly $9\times$ more than removing the entire instruction ($0.0479$ vs.\ $0.0054$), matching cross-video shifts ($0.0503$) while category profiles stay close ($0.0083$). Rather than switching global expert budgets by task category as in Explicitly Guided MoE, learned routers condition global budgets on video content and denoising stage, whereas localized instruction edits act within a largely stable sequence budget. MoE-Attention shows smaller global shifts ($0.010$ and $0.009$ vs.\ $0.001$) because it precedes cross-attention and maintains a balanced key--value pool for global queries, expressing fine-grained selectivity via attention weights $\operatorname{softmax}(L)$ rather than router-share shifts.

\paragraph{Content-sensitive routing follows object motion.}
Since router-level specialization is most pronounced in MoE-MLP, Figure~\ref{fig:expert_motion} visualizes gate weights of an illustrative content-sensitive MLP expert (Layer 14, $\sigma=0.355$). High weights coherently follow the moving person across visually distinct regions (white cap, skin, dark top) while staying sparse on dark background equipment, bright walls, and the vacated initial region (dashed), pointing to contextualized object features rather than static spatial or simple color priors.

\begin{figure}[t]
    \centering
    \includegraphics[pagebox=cropbox,width=\linewidth]{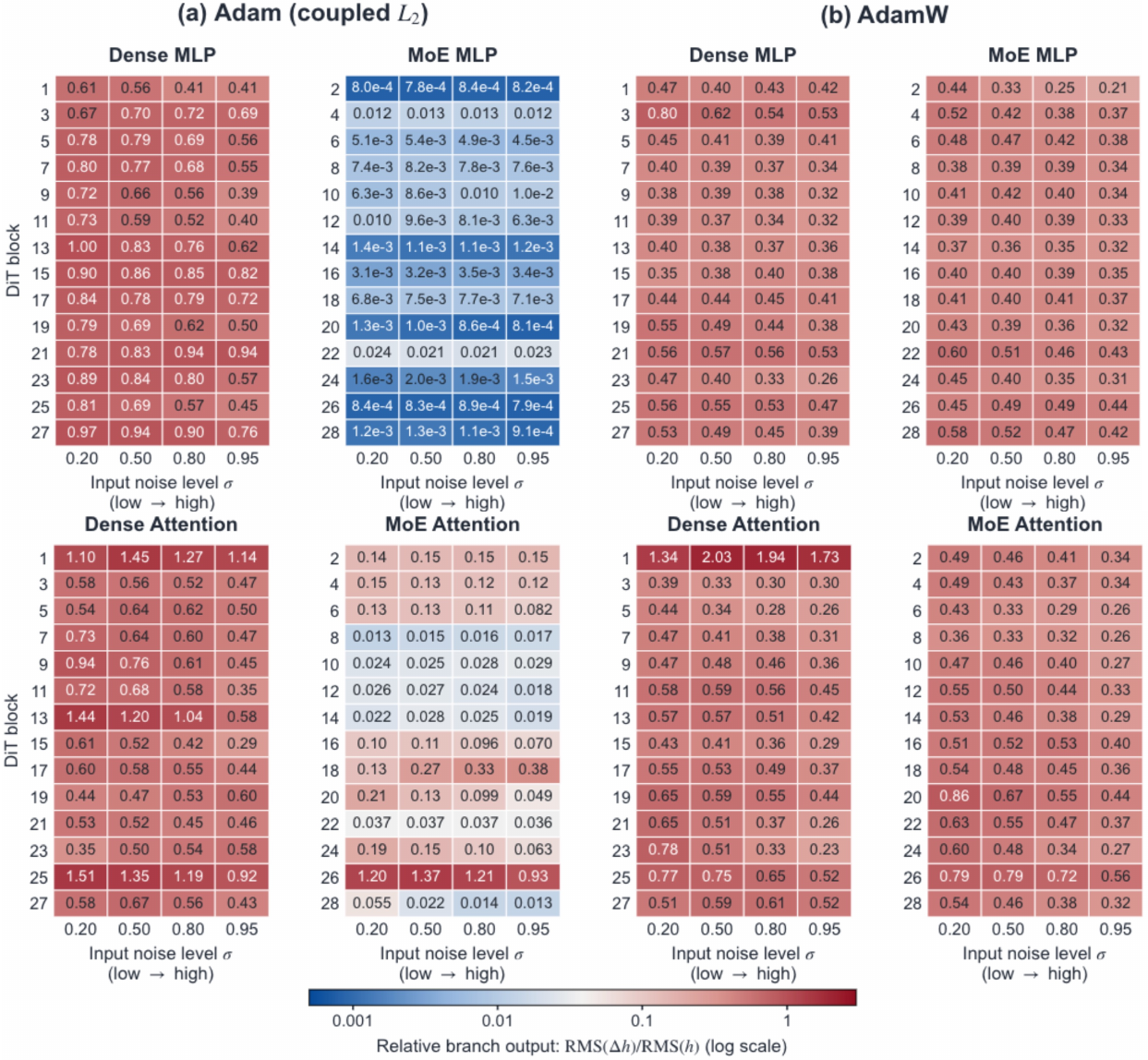}
    \caption{\textbf{Asymmetric MoE branch collapse at 56B-A8B.} MLP (top) and attention (bottom) relative outputs $c_\ell$ (sample medians) across noise levels $\sigma$ under (a) coupled-$L_2$ Adam and (b) AdamW. Coupled regularization selectively suppresses sparsely routed MoE branches while dense blocks stay active; decoupled decay restores parity.}
    \label{fig:scaling_all}
\end{figure}

\subsection{Scaling Stability}
\label{sec:scaling_stability}
\paragraph{Asymmetric expert erosion at 56B scale.}
Expanding pretrained checkpoints into a large expert pool at fixed top-$k$ creates a sharp optimization asymmetry between dense and sparsely routed branches at 56B-A8B. Figure~\ref{fig:scaling_all} measures relative branch output across noise levels $\sigma$ as $c_\ell=\operatorname{RMS}(\Delta h_\ell)/\operatorname{RMS}(h_\ell)$, where $h_\ell^{+}=h_\ell+\Delta h_\ell$ and $\operatorname{RMS}(x)=\sqrt{n^{-1}\sum_{j=1}^{n}x_j^2}$ over $n$ elements. Here, $h_\ell$ is the branch input and $\Delta h_\ell$ is the modulated residual output, including expert fusion for MoE branches. Cells report sample medians, comparing each MoE block with the preceding dense block from low to high $\sigma$.

Under Adam~\citep{adam} with coupled $L_2$ regularization (Figure~\ref{fig:scaling_all}(a)), adjacent dense blocks stay active across $\sigma$, whereas MoE branches are severely suppressed: median MoE-to-dense output ratios drop to $0.005$ for MLP and $0.12$ for attention. Checkpoint statistics confirm that pretrained MLP expert weights erode to a median $1.62\%$ of their source RMS scale, versus $52.4\%$ for dense MLP weights in the same run.

This disparity stems from task-gradient dilution under sparse routing. Every token updates dense parameters, whereas each routed expert receives task gradients only on selected tokens, scaled by $G(x)_i$. Sequential routed projections compound this shrinkage in MoE-MLP, while normalized softmax weights and a shared output projection partially buffer MoE-Attention ($0.12$ vs.\ $0.005$). Since regularization acts on all weights $W$ at every step regardless of routing, adding $\lambda W$ (coefficient $\lambda$) before moment estimation lets it \textbf{dominate the optimizer's moment estimates}: MLP expert weights and first-moment states reach a $0.99996$ median cosine similarity, producing scale-insensitive steps that erase pretrained expert weights as the pool grows.

Large-pool MoE expansion from pretrained weights is thus uniquely sensitive to moment-coupled regularization even when dense layers train normally. Decoupling weight decay ($-\eta\lambda W$ at learning rate $\eta$) from task moments via AdamW~\citep{adamw} scales shrinkage with weight magnitude while tracking sparse task gradients alone, restoring median MoE-to-dense ratios to $0.97$ for MLP and $0.93$ for attention in Figure~\ref{fig:scaling_all}(b). Diagnostic settings, evaluation protocols, and training histories appear in Appendix~\ref{app:scaling_protocol}.

\section{Conclusions}
We present VideoEditor-MoE, the first open-source MoE framework for video editing under complex instructions. By combining MoE-MLP with MoE-Attention, the framework scales to 14B-A8B and 56B-A8B variants while preserving global token interactions. Our analyses demonstrate active utilization of all experts and reveal that routing responds more strongly to sample content and diffusion noise levels than to human-defined editing categories, with spatial evidence of sensitivity to object-level semantic features. We further diagnose asymmetric pretrained expert erosion under sparse gradients when scaling to 56B. Together, these contributions provide an open-source foundation for further research on MoE architectures for video generation and editing.

\clearpage

\bibliographystyle{abbrvnat}
\bibliography{iclr2027_conference}

\clearpage
\appendix
\beginappendix

\section{Implementation Details}
\label{app:implementation}

\paragraph{Model and training.}
VideoEditor-MoE is initialized from the pretrained 5B video editing DiT of Kiwi-Edit~\citep{kiwiedit}, which builds on Wan2.2-TI2V-5B~\citep{wan22}. The backbone contains 30 blocks with hidden dimension 3072, MLP intermediate dimension 14336, and 24 attention heads of dimension 128. Training clips contain 49, 81, 121 frames at $480\times832$ resolution. The full DiT is fine-tuned, while the VLM~\citep{qwen} and VAE remain frozen. Router scores and reconstruction losses are computed in float32. Training timesteps are sampled uniformly from a 1,000-step schedule with shift 5, without additional noise-level loss weighting. Instructions are replaced with empty text with probability 0.1 while source-video conditioning is retained.

\paragraph{Optimization and inference.}
AdamW~\citep{adamw} uses separate learning rates for backbone, expert, and router parameters (Table~\ref{tab:implementation}). Expert and router learning rates decrease linearly during the first 2,000 updates and remain constant thereafter. The corrected 56B-A8B training run offloads optimizer states to CPU, uses AdamW updates, and enables ZeRO++~\citep{zero++} hpZ with partition size 8; the 14B-A8B run uses partition size 1. The corrected 56B run uses five clips per GPU and a global batch size of 120.  Outputs contain 49 frames at $480\times832$ resolution and are saved at 16 frames per second. Source videos are processed by taking their first 49 native frames, resizing to cover the target resolution, and center-cropping. The VLM's visual positional-encoding buffer was uninitialized during the 14B training run; evaluation uses corrected initialization of this buffer after training, without changing the trained DiT checkpoint.

\begin{table}[!htb]
    \centering
    \small
    \caption{Training configuration for VideoEditor-MoE-14B-A8B. Learning-rate transitions span the first 2,000 updates.}
    \label{tab:implementation}
    \begin{tabular}{ll}
        \toprule
        Setting & Value \\
        \midrule
        Clips per GPU / global batch size & $8\,/\,192$ \\
        Backbone learning rate & $10^{-5}$ \\
        Expert learning rate & $10^{-4}\rightarrow10^{-5}$ \\
        Router learning rate & $10^{-3}\rightarrow10^{-4}$ \\
        Optimizer / weight decay & AdamW / 0.01 \\
        Gradient clipping norm & 1.0 \\
        Training schedule steps / shift & $1{,}000\,/\,5$ \\
        Instruction dropout & 0.1 \\
        Motion weight / balancing coefficient & $0.1\,/\,10^{-4}$ \\
        Residual threshold & 50 \\
        \bottomrule
    \end{tabular}
\end{table}

\paragraph{Attention implementation and parameter accounting.}
Keys and values use the same selected expert indices and routing weights. Query and key normalization follow the base model; each selected key copy receives its original token's RoPE~\citep{rope} position. The $kT$ key--value candidates remain separate throughout attention. The logarithm in Equation~\ref{eq:moe_attention_output} is evaluated as $\log\max(\hat g,10^{-6})$ elementwise. If all selected expert copies produce identical keys and values and the clamp is inactive, their normalized routing weights sum to one, so Equation~\ref{eq:moe_attention_output} reduces exactly to the corresponding dense attention. This is a property of the formulation, not an assumption that the actual pretrained initialization sources are identical. The 14B and 56B counts refer to the DiT backbones; including the other components gives approximately 18.5B and 60.5B parameters. Activated parameters are counted per token and do not imply equal total memory use, latency, or attention FLOPs.

\section{Detailed Quantitative Results}
Tables~\ref{tab:openve_results}, \ref{tab:coinve_results}, and~\ref{tab:cove_results} report the scores on each benchmark. Table~\ref{tab:selected_case_results} reports their unweighted mean.

\begin{table}[!htb]
    \centering
    \caption{Quantitative comparison over OpenVE-Bench~\citep{openve}.}
    \label{tab:openve_results}
    \small
    \setlength{\tabcolsep}{3pt}
    \begin{tabular*}{\linewidth}{@{\extracolsep{\fill}}l cccc cc cc@{}}
        \toprule
        & \multicolumn{4}{c}{Existing methods} & \multicolumn{2}{c}{MoE baselines} & \multicolumn{2}{c}{Ours} \\
        \cmidrule(lr){2-5} \cmidrule(lr){6-7} \cmidrule(l){8-9}
        Metric & Kiwi-Edit & \begin{tabular}{@{}c@{}}Kiwi-Edit\\(Finetuned)\end{tabular} & \begin{tabular}{@{}c@{}}MiniMax-\\H3\end{tabular} & \begin{tabular}{@{}c@{}}CoinVE-\\Edit\end{tabular} & \begin{tabular}{@{}c@{}}Explicit-\\MoE\end{tabular} & \begin{tabular}{@{}c@{}}MoE-MLP\\(19B-A8B)\end{tabular} & 14B-A8B & 56B-A8B \\
        \midrule
        Edit Execution Rate $\uparrow$ & 0.769 & 0.837 & 0.758 & 0.875 & 0.856 & 0.939 & 0.951 & \textbf{0.977} \\
        Edit Fulfillment $\uparrow$ & 0.777 & 0.828 & 0.748 & 0.879 & 0.860 & 0.949 & 0.955 & \textbf{0.977} \\
        Video Quality $\uparrow$ & 0.670 & 0.580 & 0.523 & 0.727 & 0.716 & 0.739 & 0.750 & \textbf{0.773} \\
        FVD/1000 $\downarrow$ & 1.039 & 1.131 & 1.234 & 1.008 & 1.008 & 1.358 & 1.209 & \textbf{0.919} \\
        \bottomrule
    \end{tabular*}
\end{table}

\begin{table}[!htb]
    \centering
    \caption{Quantitative comparison over CoinVE-Bench~\citep{coinve}.}
    \label{tab:coinve_results}
    \small
    \setlength{\tabcolsep}{3pt}
    \begin{tabular*}{\linewidth}{@{\extracolsep{\fill}}l cccc cc cc@{}}
        \toprule
        & \multicolumn{4}{c}{Existing methods} & \multicolumn{2}{c}{MoE baselines} & \multicolumn{2}{c}{Ours} \\
        \cmidrule(lr){2-5} \cmidrule(lr){6-7} \cmidrule(l){8-9}
        Metric & Kiwi-Edit & \begin{tabular}{@{}c@{}}Kiwi-Edit\\(Finetuned)\end{tabular} & \begin{tabular}{@{}c@{}}MiniMax-\\H3\end{tabular} & \begin{tabular}{@{}c@{}}CoinVE-\\Edit\end{tabular} & \begin{tabular}{@{}c@{}}Explicit-\\MoE\end{tabular} & \begin{tabular}{@{}c@{}}MoE-MLP\\(19B-A8B)\end{tabular} & 14B-A8B & 56B-A8B \\
        \midrule
        Edit Execution Rate $\uparrow$ & 0.765 & 0.731 & 0.870 & 0.907 & 0.864 & 0.977 & \textbf{0.990} & \textbf{0.990} \\
        Edit Fulfillment $\uparrow$ & 0.753 & 0.731 & 0.865 & 0.902 & 0.874 & 0.983 & 0.990 & \textbf{0.993} \\
        Video Quality $\uparrow$ & 0.620 & 0.580 & 0.730 & 0.720 & 0.720 & 0.740 & 0.750 & \textbf{0.770} \\
        FVD/1000 $\downarrow$ & 1.165 & 1.129 & 1.609 & 1.062 & 1.087 & 1.170 & 1.142 & \textbf{0.949} \\
        \bottomrule
    \end{tabular*}
\end{table}

\begin{table}[!htb]
    \centering
    \caption{Quantitative comparison over CoVE-Bench~\citep{covebench}.}
    \label{tab:cove_results}
    \small
    \setlength{\tabcolsep}{3pt}
    \begin{tabular*}{\linewidth}{@{\extracolsep{\fill}}l cccc cc cc@{}}
        \toprule
        & \multicolumn{4}{c}{Existing methods} & \multicolumn{2}{c}{MoE baselines} & \multicolumn{2}{c}{Ours} \\
        \cmidrule(lr){2-5} \cmidrule(lr){6-7} \cmidrule(l){8-9}
        Metric & Kiwi-Edit & \begin{tabular}{@{}c@{}}Kiwi-Edit\\(Finetuned)\end{tabular} & \begin{tabular}{@{}c@{}}MiniMax-\\H3\end{tabular} & \begin{tabular}{@{}c@{}}CoinVE-\\Edit\end{tabular} & \begin{tabular}{@{}c@{}}Explicit-\\MoE\end{tabular} & \begin{tabular}{@{}c@{}}MoE-MLP\\(19B-A8B)\end{tabular} & 14B-A8B & 56B-A8B \\
        \midrule
        Edit Execution Rate $\uparrow$ & 0.502 & 0.551 & 0.580 & 0.556 & 0.451 & 0.515 & 0.571 & \textbf{0.591} \\
        Edit Fulfillment $\uparrow$ & 0.495 & 0.522 & 0.568 & 0.553 & 0.446 & 0.507 & 0.556 & \textbf{0.584} \\
        Video Quality $\uparrow$ & 0.486 & 0.569 & 0.542 & 0.583 & 0.514 & 0.569 & 0.611 & \textbf{0.639} \\
        FVD/1000 $\downarrow$ & 1.435 & 1.197 & 1.637 & 1.214 & 1.202 & 1.276 & 1.227 & 1.208 \\
        \bottomrule
    \end{tabular*}
\end{table}

\section{Baseline Architectures}
\label{app:baselines}

\begin{figure}[t]
    \centering
    \includegraphics[width=\linewidth]{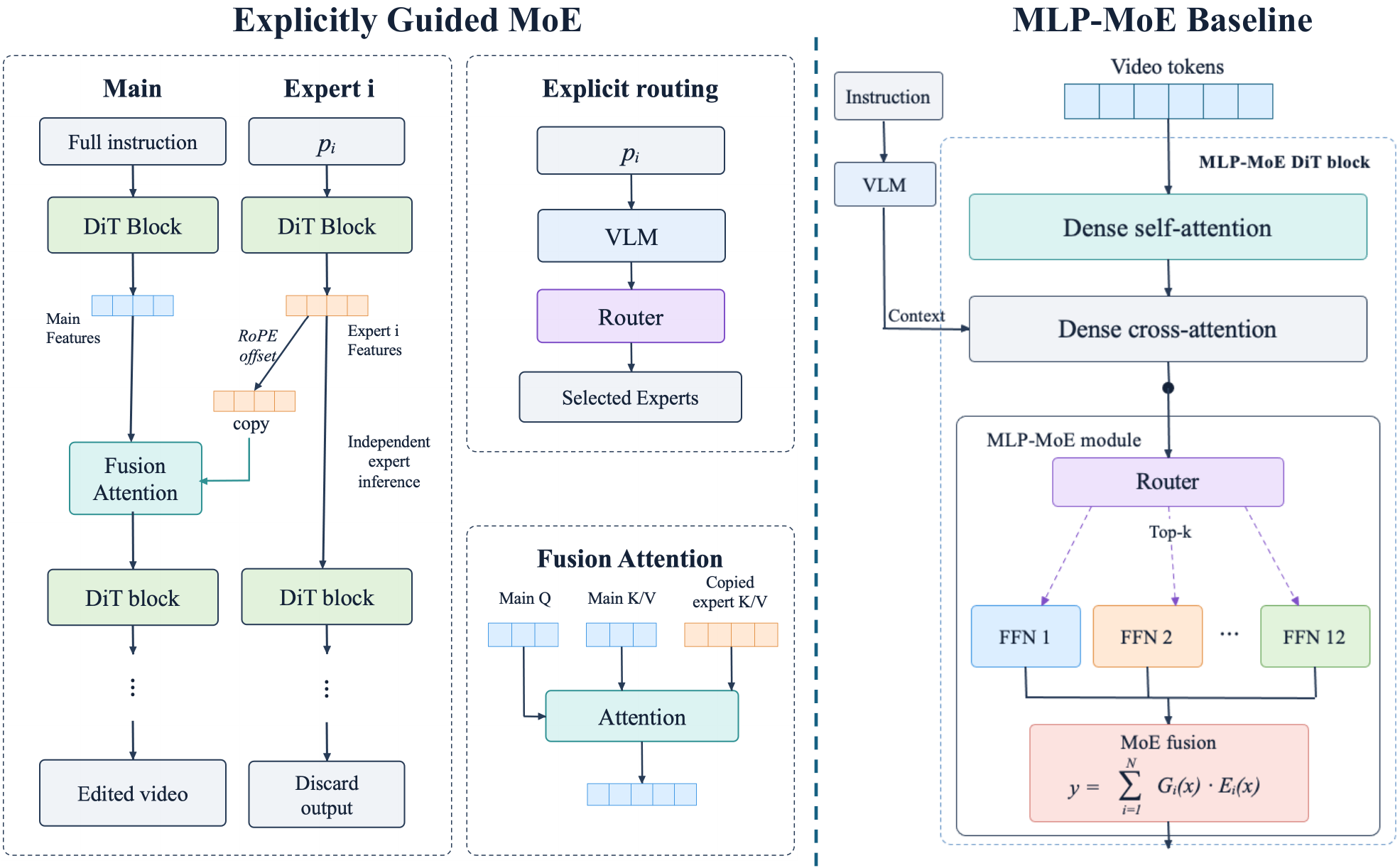}
    \caption{\textbf{Two MoE baselines.} Left: Explicitly Guided MoE sends the complete instruction to the main branch and decomposed instructions to selected experts. Fusion layers are injected into the first half of the main branch, incorporating expert features. Right: MoE-MLP routes contextualized video tokens to MLP experts while retaining dense attention.}
    \label{fig:baseline}
\end{figure}

Figure~\ref{fig:baseline} contrasts explicit task-guided expert composition with token-level MLP routing. Only the main branch produces the final video in Explicitly Guided MoE; expert branches supply features through one-way attention. MoE-MLP uses a single editing backbone with dense attention and sparse MLP layers.

\section{Training Objective}
\label{app:loss}

VideoEditor-MoE combines the flow-matching~\citep{flow_matching} reconstruction and motion objectives of the base model~\citep{kiwiedit} with an expert load-balancing loss. These terms supervise the edited video, encourage accurate temporal changes, and discourage expert underutilization, respectively.

\paragraph{Reconstruction loss.}
Given the target video latent $x_0$ and noise $\epsilon\sim\mathcal{N}(0,I)$, the flow-matching input and target velocity are
\begin{equation}
    x_\sigma=(1-\sigma)x_0+\sigma\epsilon,
    \qquad
    v^\star=\epsilon-x_0,
    \label{eq:flow_target}
\end{equation}
where $\sigma\in[0,1]$ follows the training noise schedule. The model predicts $\hat v=v_\theta(x_\sigma,c,\sigma)$, conditioned on the source video and complete instruction through $c$. Reconstruction supervision matches the predicted velocity to its target:
\begin{equation}
    \mathcal{L}_{\mathrm{rec}}
    =\mathbb{E}\!\left[
        \rho_\tau(\hat v-v^\star)
    \right],
    \qquad
    \rho_\tau(r)
    =\frac{1}{D_r}
    \sum_{a=1}^{D_r}
    r_a^2\,\mathbf{1}[|r_a|\leq\tau],
    \label{eq:reconstruction_loss}
\end{equation}
where $D_r$ is the number of elements in residual tensor $r$, and $\mathbf{1}[\cdot]$ is the indicator function. The expectation is over training examples, noise, and noise levels. The elementwise mask excludes residuals larger than $\tau=50$ to limit the influence of extreme errors; normalization still uses all $D_r$ elements. This objective learns the conditional velocity field used to generate the edited video.

\paragraph{Motion loss.}
To explicitly supervise changes between consecutive latent frames, the motion objective matches temporal differences of the predicted and target velocities:
\begin{equation}
    \mathcal{L}_{\mathrm{motion}}
    =\mathbb{E}\!\left[
        \rho_\tau\!\left(
            \Delta_f\hat v-\Delta_f v^\star
        \right)
    \right],
    \qquad
    (\Delta_f u)_f=u_{f+1}-u_f.
    \label{eq:motion_loss}
\end{equation}
Here, $f$ indexes latent frames, and the differences are computed between consecutive frames. Penalizing differences in the velocity prediction error encourages the model to reproduce the target's temporal changes.

\paragraph{Expert load balancing.}
Sparse routing can concentrate computation on a small subset of experts. Following DeepSeekMoE~\citep{deepseekmoe}, a balancing objective is computed separately for each sample and each attention or MLP router. For a sample with $T$ tokens, define
\begin{equation}
    f_i^{\ell,r}
    =\frac{N}{kT}
    \sum_{t=1}^{T}
    \mathbf{1}\!\left[
        i\in\mathcal{I}_k^{\ell,r}(t)
    \right],
    \qquad
    P_i^{\ell,r}
    =\frac{1}{T}
    \sum_{t=1}^{T}p_{t,i}^{\ell,r},
    \label{eq:balance_statistics}
\end{equation}
where $\ell\in\mathcal{M}$ indexes MoE blocks and $r\in\{\mathrm{attn},\mathrm{mlp}\}$ identifies the router. The quantity $f_i^{\ell,r}$ measures expert $i$'s selection frequency relative to uniform allocation, while $P_i^{\ell,r}$ is its average router probability. Here, $p_{t,i}^{\ell,r}$ is the probability over all $N$ experts before top-$k$ selection, as defined in Equation~\ref{eq:moe_routing}. The loss sums contributions from both routers across MoE blocks:
\begin{equation}
    \mathcal{L}_{\mathrm{bal}}
    =
    \sum_{\ell\in\mathcal{M}}
    \sum_{r\in\{\mathrm{attn},\mathrm{mlp}\}}
    \mathbb{E}_{\mathrm{sample}}
    \left[
        \sum_{i=1}^{N}
        f_i^{\ell,r}P_i^{\ell,r}
    \right].
    \label{eq:balance_loss}
\end{equation}
Coupling observed expert loads with differentiable routing probabilities discourages persistent preference for heavily used experts, without prescribing editing-category assignments.

The overall training objective is
\begin{equation}
    \mathcal{L}
    =
    (1-\lambda_m)\mathcal{L}_{\mathrm{rec}}
    +\lambda_m\mathcal{L}_{\mathrm{motion}}
    +\alpha\mathcal{L}_{\mathrm{bal}},
    \label{eq:total_loss}
\end{equation}
where $\lambda_m$ controls the contribution of temporal supervision and $\alpha$ controls the strength of expert load balancing.

The implementation uses $\lambda_m=0.1$, $\alpha=10^{-4}$, and $\tau=50$.

\section{Evaluation Protocol}
\label{app:selected_case_protocol}

\paragraph{Scoring.}
Evaluation uses Qwen3.6-27B~\citep{qwen3} with the instruction and aligned source and edited frames.

\begin{figure}[htbp]
    \centering
    \includegraphics[width=\linewidth,height=0.78\textheight,keepaspectratio]{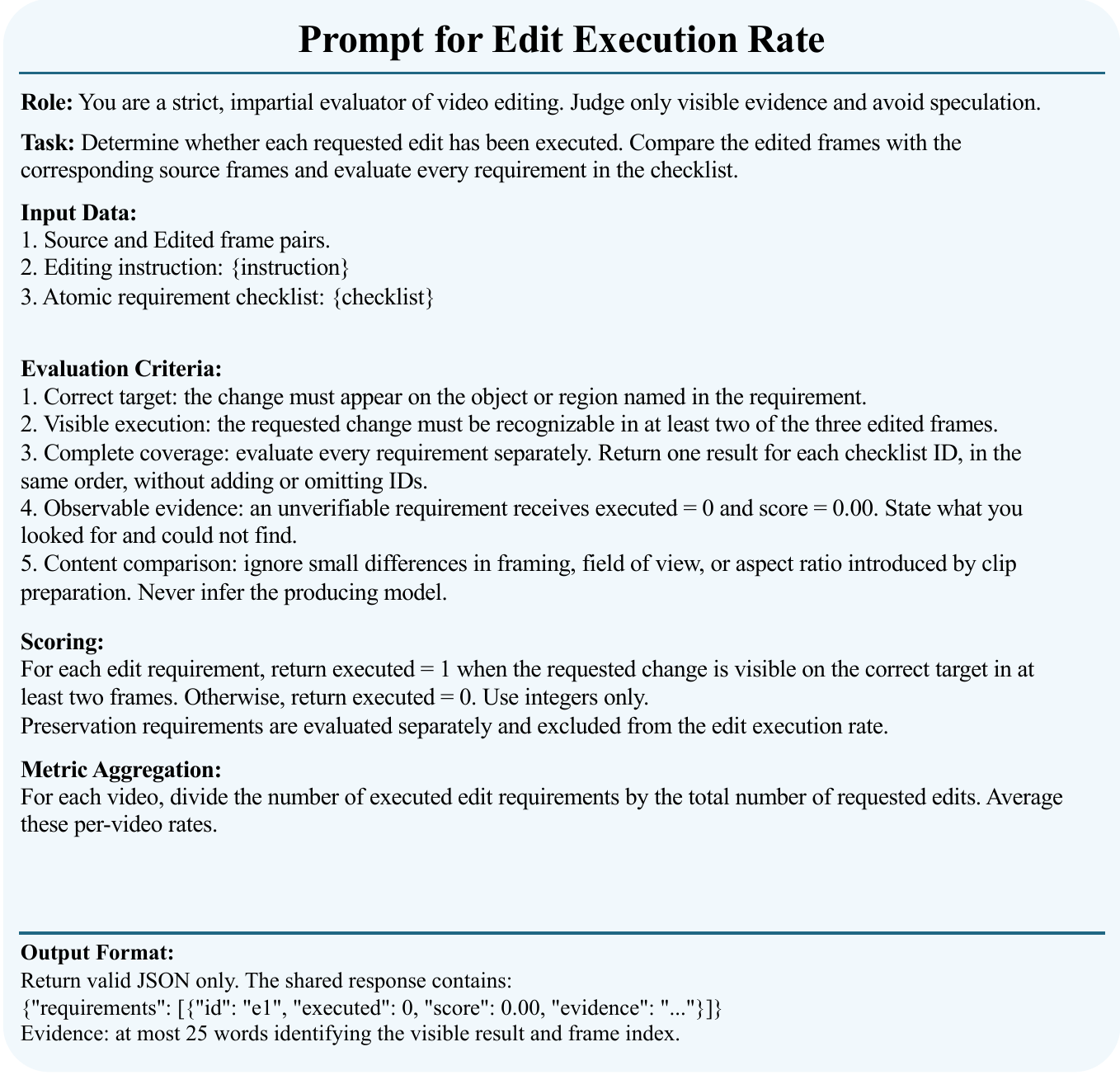}
    \caption{Evaluation criteria and output format for Edit Execution Rate.}
    \label{fig:prompt1}
\end{figure}

\begin{figure}[p]
    \centering
    \includegraphics[width=\linewidth,height=0.82\textheight,keepaspectratio]{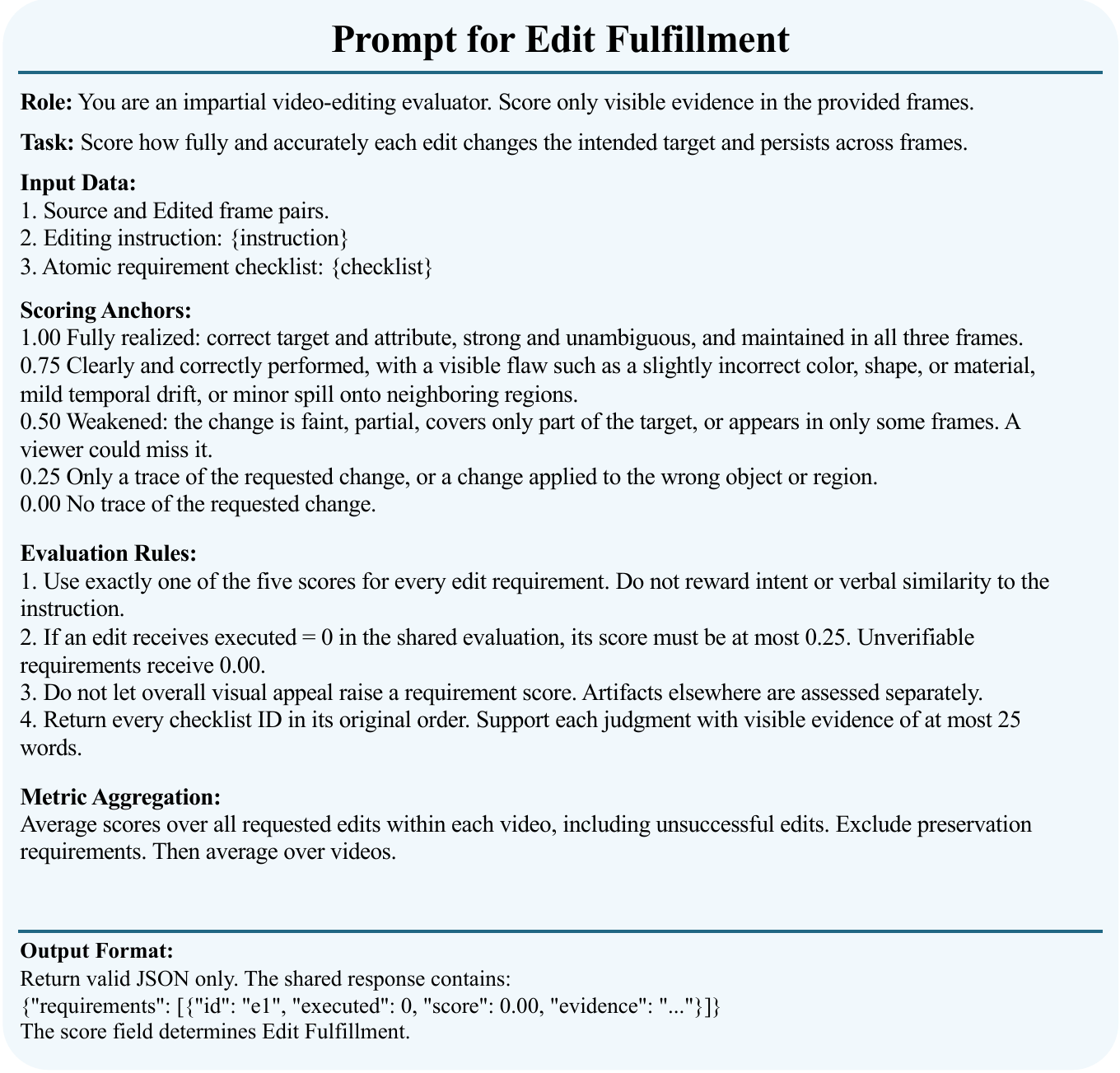}
    \caption{Evaluation criteria and output format for Edit Fulfillment.}
    \label{fig:prompt2}
\end{figure}

\begin{figure}[p]
    \centering
    \includegraphics[width=\linewidth,height=0.82\textheight,keepaspectratio]{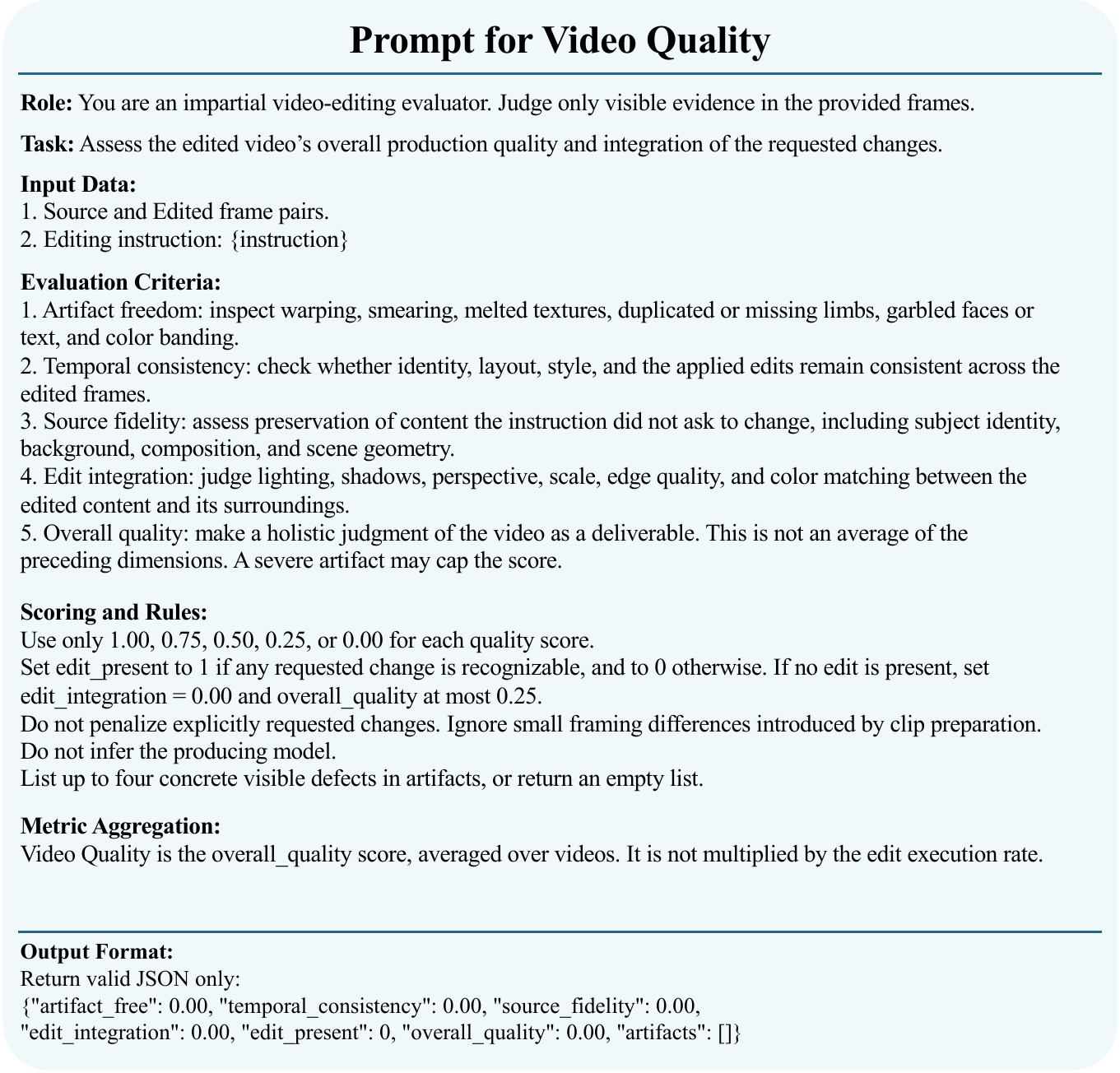}
    \caption{Evaluation criteria and output format for Video Quality.}
    \label{fig:prompt3}
\end{figure}

\paragraph{Prompt templates.}
Figures~\ref{fig:prompt1}, \ref{fig:prompt2}, and~\ref{fig:prompt3} present the criteria for execution, fulfillment, and quality. The first two panels describe the two scores obtained from the joint edit-evaluation prompt.

\section{Additional Analysis Details}
\label{app:analysis_details}

\subsection{Scaling Diagnostics}
\label{app:scaling_protocol}

The analysis compares existing 56B training runs using Adam~\citep{adam} with coupled $L_2$ regularization and AdamW~\citep{adamw}. Both use a regularization coefficient of $0.01$; the comparison is therefore a diagnostic study rather than a controlled single-variable ablation. Weight and optimizer-state statistics use checkpoints at trainer step 1,500. The Adam run has an optimizer step count of 917 following a weights-only restart, whereas the AdamW run has 1,500 optimizer steps. Moment-based statistics use the corresponding optimizer counters.

For a tensor $x$ with $n$ elements, $\operatorname{RMS}(x)=\sqrt{n^{-1}\sum_{j=1}^{n}x_j^2}$. Weight scales in Section~\ref{sec:scaling_stability} are measured as $\operatorname{RMS}(W)/\allowbreak\operatorname{RMS}(W_{\mathrm{source}})$ relative to the source checkpoints used to initialize each matrix, rather than a saved step-zero checkpoint. Reported category-level weight scales and weight-to-first-moment cosine similarities are medians of per-matrix values. Dense MLP statistics include only dense MLP weight matrices, excluding biases, normalization parameters, and other backbone modules.

\paragraph{Controlled noise-level evaluation.}
Forward evaluations compare the Adam checkpoint at trainer step 1,500 with the AdamW checkpoint at step 1,000 across cached video samples at $\sigma\in\{0.2015,\allowbreak 0.5009,\allowbreak 0.8,\allowbreak 0.95\}$; the first two values are displayed as $0.20$ and $0.50$ in Figure~\ref{fig:scaling_all}. For each sample, the source and target latents, instruction context, and one Gaussian noise realization $\epsilon$ are fixed across noise levels and runs. The input $z_\sigma=(1-\sigma)z_0+\sigma\epsilon$ mixes the clean target latent $z_0$ with this noise, and the model receives the corresponding timestep. Varying $\sigma$ changes noise strength, whereas changing the random seed would change the noise realization. These no-gradient BF16 forward passes evaluate controlled denoising conditions at fixed weights, without following a full generation trajectory. Testing several noise levels assesses whether branch-output suppression persists across the sampled conditions; it does not establish behavior at every possible noise level.

\paragraph{Branch-output measurement.}
For each attention or MLP branch, $h_\ell$ denotes the hidden state immediately before that branch's additive update, and $h_\ell^{+}$ denotes the state immediately after it. The measured output is
\begin{equation}
    h_\ell^{+}=h_\ell+\Delta h_\ell,
    \qquad
    c_\ell=\frac{\operatorname{RMS}(\Delta h_\ell)}{\operatorname{RMS}(h_\ell)}.
    \label{eq:branch_output_measurement}
\end{equation}
Here, $\Delta h_\ell$ names the final output added through the residual connection; it is neither an additional network module nor an optimizer update to the parameters. For MoE-MLP, the module first computes
\begin{equation}
    y_{\ell,t}=\sum_{i\in\mathcal I_k(x_{\ell,t})}G_i(x_{\ell,t})E_i(x_{\ell,t}),
    \label{eq:branch_mlp_fusion}
\end{equation}
where $x_{\ell,t}$ is the token representation supplied to the MLP after input normalization and modulation. The selected experts can differ for every token. Let $Y_\ell\in\mathbb R^{T\times d}$ stack the token outputs as rows, with hidden dimension $d$. $\Delta h_\ell$ is $Y_\ell$ after the DiT's output modulation; omitting that modulation gives $\Delta h_\ell=Y_\ell$. There is no parallel shared MLP expert. Attention outputs likewise include their output projection and modulation. RMS is taken over the full valid output tensor after these operations, not averaged over individual expert RMS values. Each branch uses its own immediate input: the MLP input already includes preceding self-attention and cross-attention updates.

\paragraph{Within-model comparison and aggregation.}
Dense references are the blocks retained within each MoE model, rather than blocks from a separate dense model. Using one-based block indices, the pairs are $(1,2),(3,4),\ldots,(27,28)$, with the dense block first; the final two dense blocks are excluded. Both members process their own hidden states at adjacent depths. For sample $a$, pair $j$, and noise level $\sigma$, we compute $r_{a,j,\sigma}=c_{a,2j,\sigma}/c_{a,2j-1,\sigma}$ separately for attention and MLP. Reported paired ratios are medians over samples, pairs, and noise levels, computed before any heatmap aggregation. Each heatmap cell in Figure~\ref{fig:scaling_all} reports the sample median of $c_\ell$ at one block and noise level. Cell numbers give $c_\ell$ itself, while the shared logarithmic color scale runs from blue for smaller outputs through white to red for larger outputs. White or dark numerals are chosen solely for readability against the cell background; their text color does not indicate significance, a separate category, or a different measurement. These measurements describe output amplitude relative to input amplitude, not an information fraction or editing-quality score. Layer and noise-level observations are correlated and are not treated as independent video samples.

\subsection{Expert Allocation Comparisons}
\label{app:expert_comparisons}

\paragraph{Distributions and coverage.}
For each evaluated input, layer, and router, $\mathbf s=[s_1,\ldots,s_N]$ contains the $N$ expert selection shares defined in Section~\ref{sec:expert_analysis}, with $\sum_i s_i=1$. Shares count top-$k$ selections rather than gate weights or output magnitudes; attention shares count selected key--value expert pairs. Analysis uses 14B-A8B ($N=7$, $k=3$, $L=14$ MoE layers) on noised video latents across $\sigma\in\{0.355,0.833,0.978\}$. Figure~\ref{fig:expert_analysis}(a) averages full-instruction selection shares equally across evaluated videos and noise levels, and nonzero selection counts are verified for all $14\times2\times7=196$ layer--router--expert units across every evaluated setting.

\paragraph{All-pairs sample comparison.}
At $\sigma=0.833$, let $\mathbf s_\ell^{(a)}$ denote the full-instruction share distribution of video $a$ at layer $\ell$ for a given router type across $M$ evaluated videos. Every unordered pair of distinct videos is compared once without a fixed reference:
\begin{equation}
    D_{\mathrm{sample}}=
    \frac{1}{L\binom{M}{2}}
    \sum_{\ell=1}^{L}\sum_{1\leq a<b\leq M}
    \operatorname{JS}\!\left(\mathbf s_\ell^{(a)},\mathbf s_\ell^{(b)}\right).
    \label{eq:sample_routing_comparison}
\end{equation}
Pairwise JS is computed within each layer before averaging, summarizing attention and MLP routers separately to quantify cross-video shifts in sequence-level expert budgets.

\paragraph{Category and matched-condition comparisons.}
Category-profile comparisons evaluate whether the five editing categories induce distinct global expert budgets at $\sigma=0.833$: selection shares are averaged within each applicable category per layer and router, and JS is averaged over all $\binom{5}{2}=10$ category pairs and $L$ layers. Because category-level averaging smooths individual sample variance, matched per-video controls directly test instruction and noise sensitivity without group averaging. Each matched pair fixes the source/target latents and Gaussian noise realization $\epsilon$, toggling either full versus empty instructions at fixed $\sigma=0.833$ or low versus high noise ($\sigma=0.355$ vs.\ $0.978$) under the full instruction. Control JS values are averaged within each video first and then equally across videos and layers. All JS values use base-two logarithms (in bits).

\clearpage
\section{Evaluation Prompt Templates}
\end{document}